\documentclass[letterpaper, 10 pt, conference]{ieeeconf}  

\IEEEoverridecommandlockouts                              

\usepackage{graphics} 
\usepackage{epsfig} 
\usepackage{mathptmx} 
\usepackage{times} 
\usepackage{amsmath} 
\usepackage{amssymb}  
\usepackage{booktabs}
\usepackage{algorithm} 
\usepackage{algpseudocode} 
\usepackage[font=footnotesize]{caption}
\usepackage{titlesec}
\usepackage{subcaption}
\title{\LARGE \bf
DroneDiffusion: Uncertainty-Guided Diffusion Path Planner for UAV Navigation
}

\author{Ivan Valuev, Iana Zhura, Valerii Serpiva, Miguel Altamirano Cabrera, and Dzmitry Tsetserukou 
\thanks{The authors are with the Intelligent Space Robotics Laboratory, Skolkovo Institute of Science and Technology, Moscow, Bolshoy Boulevard 30, bld.  1, 121205, Moscow, Russia.
\tt \{Ivan.Valuev, Iana.Zhura, Valerii.Serpiva, M.Altamirano, D.Tsetserukou\}@skoltech.ru}
}

\begin{document}
\bstctlcite{IEEEexample:BSTcontrol}

\maketitle
\thispagestyle{empty}
\pagestyle{empty}

\begin{abstract}
We present DroneDiffusion, a perception-aware diffusion planner for local 3-D UAV trajectory generation that explicitly models and avoids failure zones caused by lens flare, overexposure, and reflective surfaces.  A vision-language perceptual safety estimator produces a pixel-aligned reliability map that highlights regions of degraded sensing.  A denoising diffusion probabilistic model (DDPM) is conditioned on this map together with the onboard RGB image, goal point, and initial velocity to generate egocentric 3-D trajectories.  To enforce obstacle avoidance at inference time without retraining, we apply training-free CHOMP guidance over a 3-D Euclidean signed distance field.  The model is trained entirely in simulation and deployed zero-shot on a real drone.   Reliability conditioning raises the mean perceptual reliability along generated trajectories to $0.71$, compared with $0.63$ for a geometric baseline, while guidance halves the collision rate of the learned prior from $14.7\%$ to $7.8\%$.  Perceptual safety conditioning and geometric guidance are therefore complementary: treating sensor reliability as a first-class signal reduces incursions into unreliable regions while maintaining competitive navigation performance.
\end{abstract}


\section{Introduction}

Autonomous UAV navigation in cluttered indoor environments exposes onboard sensors to perceptual hazards that geometric planners routinely ignore: overexposed corridors, lens flare, transparent walls, and specular surfaces.  A planner unaware of such failures may route a drone confidently through a region where the sensor stream cannot be trusted.

Diffusion models have recently emerged as expressive trajectory generators for navigation and manipulation, offering flexible multimodal conditioning and strong generalization from data~\cite{c8,c9,c15}.  Yet existing diffusion planners condition only on geometric and semantic cues such as RGB images, goal specifications, or occupancy maps.  They provide no mechanism to reason about input reliability when the image is saturated or the depth measurements are invalid.  Vision-language tools such as CLIPSeg~\cite{c34} and AnyTraverse~\cite{c35} can detect such failure zones, but have not been integrated into trajectory-level planning.

We present a diffusion planner for 3-D UAV trajectory generation that treats a perceptual safety map as a first-class conditioning signal alongside the RGB image, goal point, and initial velocity.  A CLIPSeg-based estimator produces a pixel-aligned reliability map that penalizes overexposure, lens flare, and transparent surfaces.  A DDPM denoiser generates egocentric 4-DoF paths ($x,y,z,\psi$) of 24 waypoints at 10\,Hz conditioned on this map, naturally steering away from unreliable regions.  Training-free CHOMP guidance on a 3-D ESDF enforces geometric obstacle avoidance at inference time.  The model trains in simulation and transfers to real drone flights without fine-tuning.

The main contributions are:
\begin{itemize}
    \item A \textbf{perceptual safety map} formulation that encodes sensor failure zones as a spatial reliability prior for diffusion trajectory planning.
    \item A \textbf{conditional DDPM planner for 3-D UAV navigation}
    jointly conditioned on visual context and perceptual reliability, with sim-to-real transfer validation.
     \item A \textbf{training-free guidance module}
    enforcing obstacle avoidance and goal reaching without retraining.
\end{itemize}

\begin{figure}[t]
\centering
\includegraphics[width=1\columnwidth]{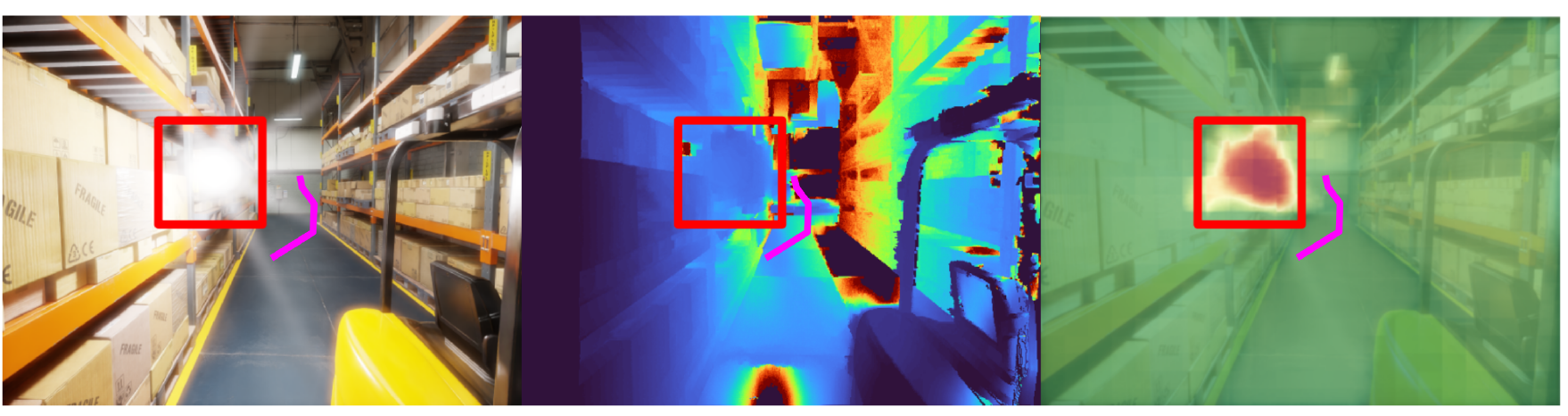}
\caption{Perception failures in indoor UAV flight.  Left: RGB observation with severe lens flare.  Middle: corresponding depth map with missing and phantom geometry in the corrupted region.  Right: our safety map flags the unreliable zone (red), guiding trajectory generation (magenta) away from it.}
\label{figurelabel0}
\vspace{-6mm}
\end{figure}

\section{Related Work}

\subsection{UAV Trajectory Planning}
Local trajectory planning for UAVs has been addressed through 
gradient-based optimization~\cite{zhou2019fastplanner}, sampling-based 
methods~\cite{c18}, and Model Predictive Control~\cite{c19}.  
These approaches generate smooth, collision-free trajectories from explicit 
geometric representations such as a signed distance field, but each query 
returns a single solution rather than a distribution over feasible behaviors.

\subsection{Diffusion Models for Robot Navigation}
Diffusion models have demonstrated strong performance as 
trajectory generators for robot navigation.  Diffuser~\cite{c9} shows that diffusion policies can plan over long horizons from offline data.  NoMaD~\cite{c15} 
extends this to goal-conditioned exploration using a masked 
diffusion policy over image observations.  DTG~\cite{c12} 
applies diffusion to mapless outdoor navigation with 
traversability-aware losses.  For UAVs specifically, CGD~\cite{c27} incorporates constraint guidance into a
diffusion planner for obstacle avoidance, while related guidance schemes have
been applied to other platforms, such as autonomous racing~\cite{c41}.

More recently, aerial diffusion planners have been extended along further
conditioning axes, including swarm path generation with generative impedance
control~\cite{batool2026impedance}, an agentic reasoning layer above the
planner~\cite{batool2026agentic}, and goals inferred from a detected person for
search and rescue~\cite{batool2026human}.  Across all of this work, the visual
input itself is treated as trustworthy.  Our work shares the guidance-based inference 
paradigm of these methods but introduces perceptual 
reliability as a conditioning modality, which none of 
them address.

\subsection{Traversability and Safety-Aware Navigation}
Traversability estimation has long been studied as a 
precondition for safe robot motion in unstructured 
environments.  Recent work has shown that 
coupling traversability directly into the diffusion 
generation process, rather than treating it as a 
post-hoc filter, yields more robust and consistent 
trajectories.  EmbodiedDiffusion~\cite{c46} demonstrates 
that safety awareness through traversability conditioning 
plays a vital role in navigation: jointly generating 
traversability maps and trajectories from a single RGB 
image achieves strong sim-to-real transfer across 
heterogeneous platforms.  Our work extends this direction 
in two respects.  Rather than encoding terrain-level 
traversability, we model sensor-level perceptual failures 
as the source of risk.  Additionally, trajectory generation 
and guidance operate in full 3-D space, addressing the geometric 
limitations of image-space planning for aerial platforms.

\subsection{Perceptual Failure Detection}
Detecting regions of degraded sensing has been studied 
in the context of robust perception pipelines.  
Vision-language models enable open-vocabulary scene 
understanding~\cite{c33}, and CLIPSeg~\cite{c34} extends 
this to dense pixel-level segmentation from text prompts.  
AnyTraverse~\cite{c35} leverages CLIPSeg to produce 
prompt-driven traversability maps for ground robots, 
covering concepts such as glass surfaces and overexposed 
regions.  These tools have not previously been used to 
condition trajectory-level planning.  We repurpose the 
AnyTraverse framework specifically to identify sensor 
failure zones and encode them as a spatial prior for 
a diffusion-based 3-D planner.

\begin{figure*}[t]
\centering
\includegraphics[width=1\textwidth]{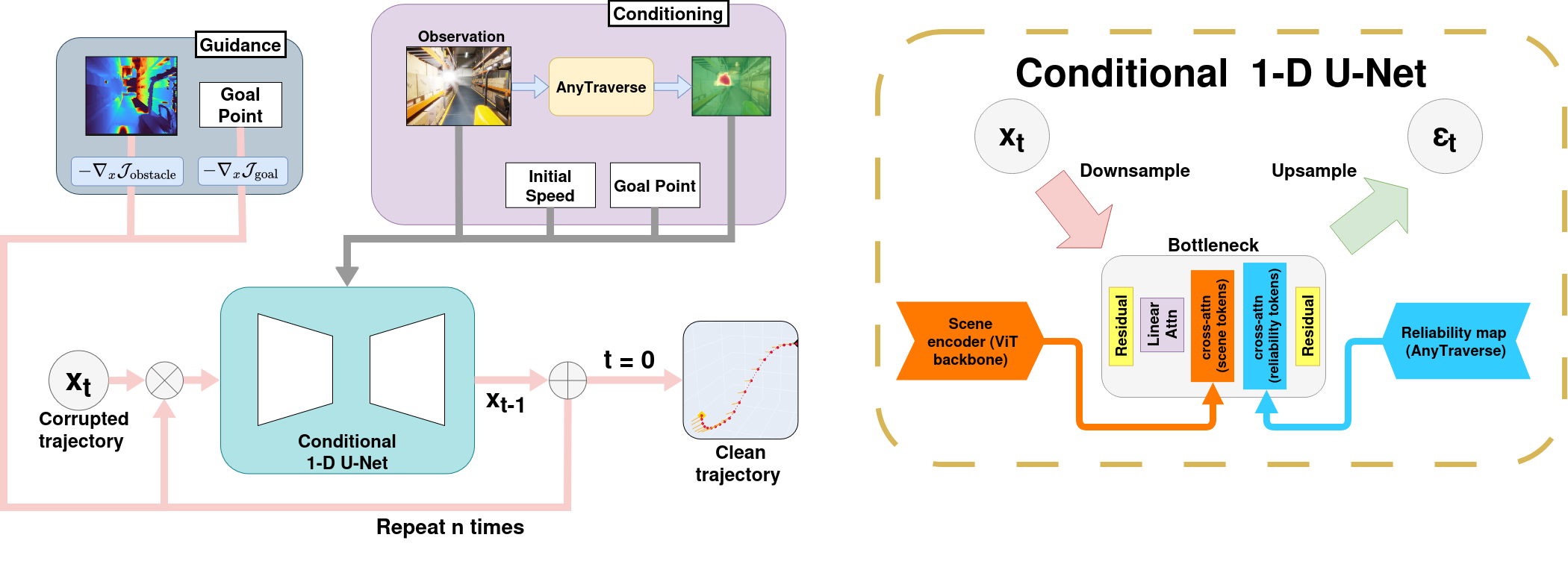}
\caption{DroneDiffusion pipeline.  Left: the corrupted trajectory $x_t$ is
denoised over $n$ reverse-diffusion steps by a conditional 1-D U-Net,
conditioned on the RGB observation, the AnyTraverse reliability map, the initial
speed and the goal point, and corrected at each step by training-free obstacle
and goal guidance.  Right: scene and reliability tokens enter the U-Net
bottleneck through two sequential cross-attention blocks, which keeps appearance
and perceptual-reliability reasoning separate.}
\label{fig:bottleneck}
\vspace{-6mm}
\end{figure*}
\section{Methods}

We propose a novel diffusion-based local UAV path-planning pipeline that accounts for both perceptual safety and geometric obstacles (Fig.~\ref{fig:bottleneck}).  It consists of three blocks: a {\bf perceptual safety} estimator that predicts reliable regions from onboard observations; a {\bf conditional diffusion planner} that generates trajectories based on the current UAV state, observation, and estimated safety map; and a {\bf cost-guidance} module that enforces collision avoidance.


The model generates an egocentric path of $H=24$ states sampled at $10 \, \mathrm{Hz}$:
\begin{equation}
x = \{(x_i,\, y_i,\, z_i,\, \sin\psi_i,\, \cos\psi_i)\}_{i=1}^{H},
\label{eq:state}
\end{equation}
where $H$ is the planning horizon, $(x_i,y_i,z_i)$ are the metric coordinates of
the $i$-th waypoint in the egocentric frame, and $\psi_i$ is its heading.   Each
state therefore has four degrees of freedom, $(x,y,z,\psi)$, stored as five
values because the heading is encoded as the unit vector
$(\sin\psi_i,\cos\psi_i)$ instead of a raw angle, which avoids angle wrapping and
preserves the smoothness of the training data.

\subsection{Perceptual Safety}

Certain scene conditions degrade the RGB and depth streams: overexposure and lens flare saturate the image and corrupt depth returns, while transparent surfaces stay invisible to the depth sensor and specular ones produce phantom geometry from reflected scene content. Because the onboard camera is the primary sensing modality for the UAV, such degradation propagates directly into planning. A planner that reads such observations at face value may confidently route the UAV through regions where the sensor data cannot be trusted. Since these failures are visible in the image itself, we exploit the ability of diffusion models to handle multimodal conditioning and feed the planner a pixel-aligned reliability map alongside the RGB observation.

This map is produced using AnyTraverse~\cite{c35}. However, it is not used to highlight traversable zones but rather potential perceptual failure zones. Built on CLIPSeg, the pipeline pools per-prompt responses with weights $\omega_i$; a weight of $0$ marks a concept as explicitly safe. The prompts for failure modes are listed in Table~\ref{tab:visibility_weights}.

\begin{table}[h]
\centering
\caption{Perceptual uncertainty prompts and pooling weights $\boldsymbol{\omega}$ used in AnyTraverse.}
\begin{tabular}{lc}
\toprule
\textbf{Prompt} & \textbf{Weight} $\omega_i$ \\
\midrule
\multicolumn{2}{l}{\emph{Transparent and specular geometry}} \\
a window                    & 1.0 \\
a glass surface             & 1.0 \\
a transparent surface       & 1.0 \\
a transparent obstacle      & 1.0 \\
a mirror                    & 1.0 \\
\midrule
\multicolumn{2}{l}{\emph{Glare and reflectivity}} \\
strong light glare          & 0.9 \\
a specular highlight        & 0.9 \\
a bright reflection         & 0.9 \\
highly reflective surfaces  & 0.9 \\
a glossy surface            & 0.8 \\
\midrule
\multicolumn{2}{l}{\emph{Negative anchors}} \\
a white wall                & 0.0 \\
\bottomrule
\end{tabular}
\label{tab:visibility_weights}
\end{table}

Since the prompts describe failure modes rather than traversable terrain, the
raw AnyTraverse output is a risk map that is high where a prompt matches and
zero elsewhere. Inverting it gives the pixel-aligned reliability map $M$:

\begin{equation}
M = 1 - \mathrm{AnyTraverse}(I;\, \boldsymbol{\omega}), \quad M \in [0,1]^{h\times w},
\label{eq:ant}
\end{equation}
where $I$ is the input RGB image, $\boldsymbol{\omega}$ is the vector of concept weights
from Table~\ref{tab:visibility_weights}, and $h$ and $w$ are the image height and
width in pixels. Each entry of $M$ scores how reliable the corresponding pixel
is, with $1$ denoting fully reliable sensing and $0$ a perceptual failure zone.
The resulting heatmap then steers trajectory generation toward reliable zones
(Fig.~\ref{figurelabel0}).


\subsection{Diffusion Planning}

Diffusion models~\cite{ho2020ddpm} are a family of generative models that iteratively generate samples from Gaussian noise.  The pipeline consists of two stages: forward and reverse.

The forward process is a Markov chain that injects Gaussian noise at each step according to a predefined schedule $\{\beta_t\}_{t=1}^{T}$:
\begin{equation}
q(x_t \mid x_{t-1}) = \mathcal{N}\!\left(x_t;\, \sqrt{1-\beta_t}\, x_{t-1},\, \beta_t \mathbf{I}\right),
\end{equation}
which admits a closed-form marginal conditioned on $x_0$:
\begin{equation}
q(x_t \mid x_0) = \mathcal{N}\!\left(x_t;\, \sqrt{\bar{\alpha}_t}\, x_0,\, (1-\bar{\alpha}_t)\,\mathbf{I}\right),
\end{equation}
with $\alpha_t = 1-\beta_t$ and $\bar{\alpha}_t = \prod_{s=1}^{t}\alpha_s$.  This allows sampling $x_t$ directly from $x_0$ as:
\begin{equation}
\label{eq:fwd}
x_t = \sqrt{\bar{\alpha}_t}\, x_0 + \sqrt{1-\bar{\alpha}_t}\, \epsilon, \quad \epsilon \sim \mathcal{N}(\mathbf{0},\mathbf{I}).
\end{equation}


The reverse process inverts (\ref{eq:fwd}) step by step.  However, without a clean sample it cannot be computed directly.  Instead, a network $\epsilon_\theta(x_t, t)$ is trained to approximate the noise $\epsilon$ injected into $x_t$ at noise level $t$, by minimizing:
\begin{equation}
\mathcal{L}_{\mathrm{diff}} = \mathbb{E}_{t,\, x_t,\, \epsilon}\!\left[\left\| \epsilon - \epsilon_\theta(x_t, t) \right\|_2^2\right].
\label{eq:loss_simple}
\end{equation}
Given this prediction, a single reverse step is performed as:
\begin{equation}
x_{t-1} = \frac{1}{\sqrt{\alpha_t}}\left(x_t - \frac{\beta_t}{\sqrt{1-\bar\alpha_t}}\,\epsilon_\theta(x_t,t)\right) + \sigma_t\,\mathbf{z},
\label{eq:reverse_step}
\end{equation}
where $\mathbf{z}\sim\mathcal{N}(\mathbf{0},\mathbf{I})$ if $t>1$, else $\mathbf{z}=\mathbf{0}$.  
Generation starts from $x_T \sim \mathcal{N}(\mathbf{0}, \mathbf{I})$ and applies (\ref{eq:reverse_step}) iteratively for $t = T, T{-}1, \dots, 1$.

The formulation above describes unconditional generation, in which samples are drawn from the learned data distribution $p_\theta(x_0)$ without any external control.  For path planning, generated samples must be consistent with the current task and environment.  This is achieved by conditioning the model on an additional signal $c$, yielding $\epsilon_\theta(x_t, t, c)$.  The resulting architecture is shown in Fig.~\ref{fig:bottleneck}, and the conditioning signals used for planning are:
\begin{itemize}
    \item Initial speed for start dynamics;
    \item Goal point for goal pursuit;
    \item RGB image for global context;
    \item Safety map for highlighting unreliable regions.
\end{itemize}

The denoiser is implemented as a 1-D U-Net with three resolution levels, chosen for its low inference latency given the short planning horizon.  Scalar and vector conditions (initial speed, goal) are injected via Feature-wise Linear Modulation (FiLM)~\cite{perez2018film} in every residual block (in stages of downsampling, bottleneck, and upsampling), while spatially structured inputs are injected via two sequential cross-attention blocks at the U-Net bottleneck: frozen CLIP ViT-B/16~\cite{c33} patch tokens from the RGB image provide scene-level semantic context, and patch tokens from a lightweight safety-map encoder provide the perceptual reliability signal (Fig.~\ref{fig:bottleneck}).  Using two independent cross-attention blocks allows the model to separately reason about scene appearance and perceptual safety without conflating the two signals.

To further shape the generated trajectory, several auxiliary losses are applied to the estimated clean sample $\hat{x}_0$, reconstructed from the noisy sample and predicted noise.  The reconstruction loss $\mathcal{L}_{\mathrm{rec}}$ is the mean squared error between $\hat{x}_0$ and the ground-truth trajectory and reinforces the ability of the network to recover the clean sample.   The smoothness loss $\mathcal{L}_{\mathrm{sm}}$ suppresses jerky motion through the squared second finite difference of the waypoint sequence, which acts as an acceleration proxy.   The origin-anchor loss $\mathcal{L}_{\mathrm{or}}$ keeps the trajectory egocentric by pulling the first state to the UAV-frame origin $[0,0,0,0,1]$.   The speed-consistency loss $\mathcal{L}_{\mathrm{sp}}$ matches the finite-difference displacement between the first two waypoints to the observed initial velocity $v_0$ scaled by the sampling interval.   The yaw-geometry loss $\mathcal{L}_{\mathrm{yaw}}$ keeps the predicted $(\sin\psi,\cos\psi)$ pair on the unit circle and aligns the heading with both the ground-truth yaw and the direction of motion along moving segments.

The perceptual-safety loss $\mathcal{L}_{\mathrm{reliab}}$ couples the trajectory to the reliability map $M$.   Each predicted waypoint is projected into the image plane through the camera intrinsics, and its reliability $s_i \in [0,1]$ is sampled from $M$ as the mean over a perspective-aware $n\times n$ kernel matching the projected UAV footprint at the waypoint depth.   Reliability is treated as a soft constraint, so each in-frame waypoint receives the penalty: \begin{equation}
  \rho(s_i)=\tau\,\mathrm{softplus}\!\left(\frac{s_{\mathrm{thr}}-s_i}{\tau}\right),
  \label{eq:reliab_penalty}
\end{equation}
where $s_{\mathrm{thr}}$ is the reliability threshold and $\tau$ is the temperature.
Waypoints above $s_{\mathrm{thr}}$ incur a vanishing penalty, while those below
it are penalized almost linearly in the deficit.   The full term averages this
penalty over the in-frame waypoints, adds a conditional tail contribution over
the least reliable fraction so that a few dangerous states are not masked by an
otherwise safe path, and includes a smooth field-of-view penalty
$\rho_{\mathrm{fov}}$ that pulls back projections leaving the frame or falling
behind the camera:
\begin{equation}
  \mathcal{L}_{\mathrm{reliab}}=
  \frac{\overline{\rho}+w_{t}\,\rho_{\mathrm{tail}}}{1+w_{t}}
  +w_{\mathrm{fov}}\,\rho_{\mathrm{fov}}.
  \label{eq:reliab_total}
\end{equation}

Combined with the main diffusion loss $\mathcal{L}_{\mathrm{diff}}$~(\ref{eq:loss_simple}), the complete training objective is a weighted sum of the losses.
The auxiliary terms are applied only to the samples drawn
from the lower $30\%$ of the noise schedule, where the clean estimate is
accurate enough to supervise trajectory geometry.

\subsection{Guidance}
\label{sec:guidance}

The denoiser provides a trajectory prior but does not guarantee collision
avoidance in unseen scenes.   Collision constraints are introduced through
training-free CHOMP guidance~\cite{ratliff2009chomp} on the estimated clean manifold
sample (ECM)~\cite{wang2024ecm}:
\begin{equation}
  \hat{x}_0 =
  \frac{x_t-\sqrt{1-\bar{\alpha}_t}\,
  \epsilon_\theta(x_t,t,c)}
       {\sqrt{\bar{\alpha}_t}}.
  \label{eq:guidance_x0}
\end{equation}
Its position channels define the trajectory
\(P=[p_1,\ldots,p_H]^T\), where \(p_i\in\mathbb{R}^3\) is the \(i\)-th waypoint.
Guidance corrects positions only; the heading channels remain as produced by the
denoiser.  Collision geometry is represented by a 3-D
Euclidean Signed Distance Field (ESDF)~\cite{oleynikova2017voxblox} \(\phi(p)\) derived from the
current depth point cloud.   The field is positive in free space, zero at an
obstacle surface, and negative inside an obstacle.   With the UAV radius and
safety margin represented by $r_{\mathrm{clr}}$, the signed clearance is:
\begin{equation}
  d_i=\phi(p_i)-r_{\mathrm{clr}}.
  \label{eq:chomp_clearance}
\end{equation}
The CHOMP obstacle potential is:
\begin{equation}
  c(d_i)=
  \begin{cases}
    0, & d_i\geq\eta,\\[2pt]
    \dfrac{(d_i-\eta)^2}{2\eta}, & 0<d_i<\eta,\\[6pt]
    -d_i+\dfrac{\eta}{2}, & d_i\leq 0,
  \end{cases}
  \qquad
  C_{\mathrm{obs}}(P)=\sum_{i=1}^{H}c(d_i),
  \label{eq:chomp_cost}
\end{equation}
where \(\eta\) is the obstacle influence distance.  The cost is zero beyond \(\eta\), quadratic in the near-obstacle free-space region, and linear inside an obstacle.   The complete objective also includes smoothness, speed, and goal terms:
\begin{equation}
\begin{split}
  C(P)={}&s_{\mathrm{obs}}C_{\mathrm{obs}}(P)+\lambda_s C_{\mathrm{smooth}}(P)\\
         &+\lambda_v C_{\mathrm{spd}}(P)+\lambda_g C_{\mathrm{goal}}(P).
\end{split}
\label{eq:guidance_total}
\end{equation}
Directional weighting of the metric-space gradient \(G=\nabla_P C(P)\) is
defined by the anisotropy vector \(A=[a_x,a_y,a_z]\):
\begin{equation}
  \widetilde{G}=G\odot A.
  \label{eq:guidance_anisotropy}
\end{equation}
Since guidance acts on positions alone, $A$ is three-dimensional, and its
coefficients independently determine the relative contribution of forward,
lateral, and vertical corrections.
Independent waypoint gradients may produce sharp or alternating corrections.
Temporal coupling is represented by:
\begin{equation}
  \Lambda=\rho_0I+\rho_2D_2^T D_2+\rho_3D_3^T D_3,
  \label{eq:covariant_metric}
\end{equation}
where \(D_2\) and \(D_3\) are the second- and third-order finite-difference
operators.   The inverse metric gives the smoothed gradient
\(\overline{G}=\Lambda^{-1}\widetilde{G}\), coupling each correction to neighboring
waypoints.   At least the first waypoint remains unchanged to preserve the
trajectory origin.   For warm-started generation, more initial waypoints can
be fixed to maintain temporal consistency with the previous plan.   If a
correction exceeds a predefined metric limit, all waypoint corrections are
reduced by the same factor.   This common scaling bounds each guidance step
while preserving the relative shape of the smoothed trajectory.   The
resulting path \(P^*\) defines the guided clean estimate \(\hat{x}_0^*\), and
the next reverse state retains its effective noise:
\begin{align}
  \widetilde{\epsilon}_t=
  \frac{x_t-\sqrt{\bar{\alpha}_t}\,\hat{x}_0^*}
       {\sqrt{1-\bar{\alpha}_t}}, \qquad
  x_{t'}=
  \sqrt{\bar{\alpha}_{t'}}\,\hat{x}_0^*
  +\sqrt{1-\bar{\alpha}_{t'}}\,\widetilde{\epsilon}_t.
  \label{eq:guided_transition}
\end{align}

The goal term $\lambda_g C_{\mathrm{goal}}$ in (\ref{eq:guidance_total}) provides goal attraction.  It is defined as:
\begin{equation}C_{\mathrm{goal}}(P)=\lVert p_H-g\rVert^2,\label{eq:goal_cost}
\end{equation}
which penalizes the distance between the trajectory endpoint $p_H$ and the local goal $g$.  Its gradient acts only on the terminal waypoint, but the inverse metric $\Lambda^{-1}$, derived from~(\ref{eq:covariant_metric}), distributes the correction over the whole trajectory, so the path bends toward the goal coherently instead of forming a sharp terminal kink.  Because the obstacle weight $s_{\mathrm{obs}}$ dominates $\lambda_g$, collision avoidance retains priority over goal attraction.

\section{Evaluation}
\label{sec:experiments}

\subsection{Dataset and Training}

The diffusion planner was trained on synthetic warehouse data generated in NVIDIA Isaac Sim~\cite{nvidia_isaacsim} with Pegasus~\cite{jacinto2024pegasus}.  The dataset comprises $\sim5{,}000$ RGB-D observation frames across eight warehouse scenes, expanded to $\sim20{,}000$ trajectory samples.
Each sample contains an RGB observation, a pixel-aligned perceptual reliability map from AnyTraverse inference, an initial velocity, a goal point, and an egocentric trajectory of $H$ states, where each state is defined by $[x,y,z,\sin\psi,\cos\psi]$.   The training-validation split was approximately 5:1.  It was performed at the environment level, so that frames from the same or a similar point of view never appeared in both training and validation.  The trajectory and velocity normalization ranges were estimated from the training split only.

The denoiser is the 1-D U-Net described above, conditioned on frozen CLIP ViT-B/16 tokens and a lightweight safety-map embedding.   It was trained with AdamW under FP16 mixed precision for $100$ epochs at a learning rate of $10^{-4}$ and batch size $128$.   The forward process used $1000$ diffusion steps with a squared-cosine schedule and epsilon prediction.   The reliability condition was dropped with probability $0.10$ so that the planner remains usable when no safety map is available.
\subsection{Hardware and Sensing}
Training and inference ran on an NVIDIA RTX 4070 Super (12 GB).  Onboard sensing used an Intel RealSense D455 providing $640\times480$ RGB-D, from which the depth stream feeds the ESDF.  The ground server and UAV operated within the same ROS 2 domain.

\subsection{Inference and Guidance Configuration}

The sampling, refinement, and geometry settings are summarized in Table~\ref{tab:inference_configuration}.   The obstacle coefficient remains constant during refinement; no timestep-dependent strength schedule is used.

\begin{table}[htbp]
\centering
\caption{Diffusion and geometric-guidance configuration.}
\label{tab:inference_configuration}
\footnotesize
\setlength{\tabcolsep}{3.5pt}
\begin{tabular}{@{}lc|lc@{}}
\toprule
\textbf{Parameter} & \textbf{Value} &
\textbf{Parameter} & \textbf{Value} \\
\midrule
Cold reverse steps & 500 & Refinement start $t$ & 300 \\
Refinement steps & 11 & ESDF voxel size & $0.05\,\mathrm{m}$ \\
UAV radius & $0.25\,\mathrm{m}$ & Safety margin & $0.10\,\mathrm{m}$ \\
CHOMP $\eta$ & $0.7\,\mathrm{m}$ & Obstacle coefficient & 0.75 \\
Maximum correction & $0.08\,\mathrm{m}$ & Anisotropy $A$ & $[0.8,1.0,1.0]$ \\
$\rho_0,\rho_2,\rho_3$ & $1.0,1.0,0.2$ & Goal weight $\lambda_g$ & 0.25 \\
Smoothness $\lambda_s$ & 1.0 & Speed $\lambda_v$ & 0.5 \\
\bottomrule
\end{tabular}
\end{table}

The local ESDF covers $x\in[-0.1,5.0]$, $y\in[-3.0,3.0]$, and
$z\in[-1.5,2.0]\,\mathrm{m}$ in the egocentric FLU frame.

\subsection{Evaluation Protocol and Metrics}

Evaluation observations came from a controlled drone flight in which trajectories were planned online by the proposed method from the onboard RGB-D stream; the reported statistics were obtained from four additional seeds replayed offline on these observations. While planning uses only the partial onboard view, collisions are verified against a precise 3-D scan of the entire arena that serves as ground truth. Each generated trajectory is transformed from the local FLU frame into the global map frame using the corresponding camera pose, sampled every $0.02\,\mathrm{m}$, and queried against this scan within the camera field of view; a sample is colliding when its nearest-map distance is below the collision-body radius in Table~\ref{tab:inference_configuration}.

The following planner variants are evaluated on identical observations, goals, and random seeds (for diffusion):
\begin{itemize}
    \item \textbf{Full (Ours)}: warm-started generation with ESDF guidance and
    safety-map conditioning;
    \item \textbf{Without guidance}: the same planner with geometric guidance
    disabled;
    \item \textbf{Without guidance and reliability}: guidance disabled and the safety map
    replaced by an all-reliable map, so perceptual conditioning is also removed;
    \item \textbf{Artificial Potential Field (APF)}~\cite{khatib1986apf}: a
    geometric non-diffusion baseline evaluated with the same collision model.
\end{itemize}

Collision rate is the fraction of generated trajectories containing at least
one colliding interpolated sample, and minimum clearance is the smallest
nearest-map distance along a trajectory.   Goal error is the Euclidean distance
between the final waypoint and the specified local goal.   Reliability is
evaluated by projecting waypoints onto the reliability map: the mean and the
lower-tail (worst-case) reliability are reported, and the unsafe-waypoint rate
is the fraction of projected waypoints below a reliability threshold ($0.2$).
Jerk is measured from finite differences of the waypoint sequence, so lower values indicate smoother motion.

\subsection{Quantitative Results}
Table~\ref{tab:trajectory_results} reports the ablation results over five diffusion seeds.  The learned planner provides a useful prior for safe and perceptually reliable trajectory generation.  Adding geometric guidance reduces the collision rate from $14.71\%$ to $7.77\%$, while leaving perceptual reliability approximately unchanged.  This improvement comes at the cost of increased goal error and higher jerk, despite including goal and smoothness terms in the guidance objective.  The APF baseline produces shorter and smoother geometric paths but does not account for perceptual reliability.  Overall, DroneDiffusion improves perceptual safety relative to the geometric baseline while remaining competitive in collision avoidance.


\begin{table}[htbp]
\centering
\caption{Trajectory-generation results on real drone observations, averaged over
five seeds (mean and std across seeds, APF is deterministic).  Best per row in bold.}
\label{tab:trajectory_results}
\scriptsize
\setlength{\tabcolsep}{1.3pt}
\renewcommand{\arraystretch}{1.05}
\resizebox{0.96\columnwidth}{!}{%
\begin{tabular}{@{}lcccc@{}}
\toprule
\textbf{Metric} &
\shortstack{\textbf{Full}\\\textbf{(Ours)}} &
\shortstack{\textbf{w/o}\\\textbf{Guid.}} &
\shortstack{\textbf{w/o}\\\textbf{Guid. and Reliab.}} &
\textbf{APF} \\
\midrule
Collision (\%) $\downarrow$ &
$7.770 \pm 1.150$ &
$14.710 \pm 2.270$ &
$13.320 \pm 1.320$ &
$\mathbf{6.650 \pm 0.000}$ \\
Min. clearance (m) $\uparrow$ &
$\mathbf{0.618 \pm 0.013}$ &
$0.602 \pm 0.014$ &
$0.610 \pm 0.016$ &
$0.613 \pm 0.000$ \\
Unsafe WP (\%) $\downarrow$ &
$\mathbf{12.890 \pm 1.550}$ &
$14.590 \pm 1.780$ &
$16.590 \pm 1.430$ &
$20.520 \pm 0.000$ \\
Mean reliability $\uparrow$ &
$0.714 \pm 0.016$ &
$\mathbf{0.730 \pm 0.013}$ &
$0.689 \pm 0.015$ &
$0.626 \pm 0.000$ \\
Tail reliability $\uparrow$ &
$0.522 \pm 0.033$ &
$\mathbf{0.538 \pm 0.005}$ &
$0.518 \pm 0.018$ &
$0.390 \pm 0.000$ \\
Goal error (m) $\downarrow$ &
$0.723 \pm 0.012$ &
$0.610 \pm 0.011$ &
$0.666 \pm 0.014$ &
$\mathbf{0.601 \pm 0.000}$ \\
Path length (m) $\downarrow$&
$3.352 \pm 0.014$ &
$3.042 \pm 0.005$ &
$3.043 \pm 0.005$ &
$\mathbf{2.687 \pm 0.000}$ \\
Jerk $\downarrow$ &
$0.0690 \pm 0.0008$ &
$0.0150 \pm 0.0002$ &
$0.0150 \pm 0.0002$ &
$\mathbf{0.0050 \pm 0.0000}$ \\
\bottomrule
\end{tabular}
}
\end{table}

\subsection{Runtime Evaluation}

Table~\ref{tab:runtime_breakdown} shows that AnyTraverse dominates runtime;
the complete-pipeline mean corresponds to a trajectory-generation rate of
approximately $5.51\,\mathrm{Hz}$.  Reliability reasoning is the main latency bottleneck.

\begin{table}[htbp]
\centering
\caption{Runtime breakdown on an NVIDIA RTX 4070 Super (12\,GB).}
\label{tab:runtime_breakdown}
\scriptsize
\setlength{\tabcolsep}{1.5pt}
\begin{tabular}{@{}lrrrr@{}}
\toprule
\textbf{Stage} & \textbf{Mean} & \textbf{Min.} &
\textbf{Max.} & \textbf{Share} \\
\midrule
ESDF & 18.38 & 16.84 & 21.17 & 10.13\% \\
AnyTraverse & 122.14 & 117.13 & 125.23 & 67.35\% \\
SceneEncoder & 5.23 & 5.06 & 5.78 & 2.88\% \\
Generation + guidance (11 steps) & 35.61 & 33.28 & 38.35 & 19.63\% \\
\midrule

\textbf{Full pipeline (warm start)} & \textbf{181.36} & \textbf{173.87} &
\textbf{186.21} & \textbf{100\%} \\
\bottomrule
\end{tabular}
\vspace{1mm}

\parbox{0.98\columnwidth}{\scriptsize Time dimension is in milliseconds.}
\end{table}

\section{Conclusion and Future Work}

We presented a perception-aware diffusion planner for safe UAV trajectory generation in indoor environments.  The key idea is to condition a lightweight 1-D U-Net denoiser jointly on visual scene context and a dense perceptual reliability map and to couple it with training-free ESDF-based cost guidance for obstacle avoidance.  This design lets the planner reason simultaneously about what is geometrically free and what is perceptually trustworthy, a distinction that classical occupancy-based planners do not make.

Ablation experiments confirm that both components contribute.  Safety-map conditioning internalizes general safety into the prior, so that even with the map removed at inference the planner stays more perceptually reliable than a deterministic geometric baseline.  CHOMP-style guidance roughly halves the collision rate on unseen obstacle configurations, at the cost of a moderate decrease in trajectory quality. More broadly, the results suggest that explicitly reasoning about sensor reliability, rather than treating all visual input as uniformly trustworthy, is a promising direction at the intersection of diffusion-based planning and perception-aware robotics.

Several limitations remain.  The training data is entirely synthetic and limited to warehouse scenes, and real-world evaluation was conducted only in a controlled environment; deployment in uncontrolled, visually diverse settings remains untested.  End-to-end latency is dominated not by the diffusion model but by the AnyTraverse reliability estimator, which accounts for about two-thirds of the pipeline latency.  Cold-start generation is also expensive: it relies on a 500-step reverse chain, which is reduced to a guided-refinement schedule to keep replanning tractable.  Finally, the reliability signal is static per frame and does not model dynamic obstacles or rapidly changing illumination.

Future work targets three directions: (i) attacking latency at its true bottleneck by distilling or replacing the AnyTraverse estimator, together with fast sampling via consistency distillation or flow matching to bring end-to-end latency below 100 ms; (ii) replacing the static reliability map with direct per-frame uncertainty estimation; and (iii) extending the goal-conditioned local planner toward long-horizon operation, including goal-free exploration in which the planner navigates and discovers unmapped environments without a specified target.

\section*{Acknowledgment}
Research reported in this publication was financially supported by the RSF grant No.  24-41-02039.

\addtolength{\textheight}{-12cm}   









\bibliographystyle{IEEEtran}
\bibliography{bib}

\end{document}